\documentclass[journal]{IEEEtran}
\usepackage{amsmath}
\usepackage{url}
\usepackage{booktabs}
\usepackage{graphicx}
\usepackage{subcaption}

\usepackage[pagebackref=true,breaklinks=true,letterpaper=true,colorlinks,bookmarks=false]{hyperref}
\newcommand{\kwang}[1]{\textcolor[rgb]{0,0,0}{#1}}
\newcommand{\pxj}[1]{\textcolor[rgb]{0,0,0}{#1}}
\newcommand{\rpxj}[1]{\textcolor[rgb]{0,0,0}{#1}}

\newcommand{\peng}[1]{\textcolor[rgb]{0,0,0}{#1}}
\begin{document}
%
\title{Region Attention Networks for Pose and Occlusion\\ Robust Facial Expression Recognition}
%
%
%

\author{Kai Wang,
        Xiaojiang Peng,
        Jianfei Yang,
        Debin Meng,
        and~Yu Qiao,~\IEEEmembership{Senior~Member,~IEEE}
 \thanks{Kai Wang and Xiaojiang Peng are equally-contributted authors.}
\thanks{Kai Wang, Xiaojiang Peng, Debin Meng and Yu Qiao are with Shenzhen Institutes of Advanced Technology, Chinese Academy of Sciences, Shenzhen, China.}
\thanks{Jianfei Yang is with School of Electrical and Electronic Engineering, Nanyang Technological University, Singapore}
\thanks{Corresponding Author: Yu Qiao(yu.qiao@siat.ac.cn)}}

%
%

\markboth{Journal of \LaTeX\ Class Files,~Vol.~14, No.~8, August~2015}%
{Shell \MakeLowercase{\textit{et al.}}: Bare Demo of IEEEtran.cls for IEEE Journals}
%



\maketitle

\begin{abstract}
Occlusion and pose variations, which can change facial appearance significantly, are two major obstacles for automatic Facial Expression Recognition (FER). Though automatic FER has made substantial progresses in the past few decades, occlusion-robust and pose-invariant issues of FER have received relatively less attention, especially in real-world scenarios. \rpxj{This paper addresses the real-world pose and occlusion robust FER problem in the following aspects. First, to stimulate the research of FER under real-world occlusions and variant poses, we annotate  several in-the-wild FER datasets with pose and occlusion attributes for the community}. Second, we propose a novel Region Attention Network (RAN), to adaptively capture the importance of facial regions for occlusion and pose variant FER. The RAN aggregates and embeds varied number of region features produced by a backbone convolutional neural network into a compact fixed-length representation. Last, inspired by the fact that facial expressions are mainly defined by facial action units, we propose a region biased loss to encourage high attention weights for the most important regions. We validate our RAN and region biased loss on both our built test datasets and four popular datasets: FERPlus, AffectNet, RAF-DB, and SFEW. Extensive experiments show that our RAN and region biased loss largely improve the performance of FER with occlusion and variant pose. Our method also achieves state-of-the-art results on FERPlus, AffectNet, RAF-DB, and SFEW. Code and the collected test data will be publicly available.
\end{abstract}

\section{Introduction}
\label{intro}
\IEEEPARstart{F}acial expressions play important roles in daily human-human communication. Automatic facial expression analysis is an important area of artificial intelligence. Due to its potential applications in various fields, such as intelligent tutoring systems, service robots, driver fatigue monitoring, Facial Expression Recognition (FER) has attracted increasing attention in the computer vision community recently~\cite{jung2015joint,barsoum2016training,fabian2016emotionet,mollahosseini2017affectnet,li2017reliable,dhall2018emotiw}.
The main challenges of FER come from illumination variation, occlusions, variant poses, identity bias, insufficient qualitative data, etc. 

Occlusions and variant poses are two major problems in the field of face analysis since \pxj{they} lead to significant change of facial appearance. These issues have received wide interest in face identity recognition~\cite{ding2016comprehensive,liao2013partial}, however, less attention has been paid in real-world FER partly due to the lack of a facial expression dataset with occlusion and pose annotations.

Earlier works mainly investigate the effects of occlusion for FER systems with partial artificially-masked faces collected in a controlled laboratory environment. \pxj{Boucher and Ekman~\cite{boucher1975facial} investigate facial parts to understand which are most important regions for human perception by occluding key parts.} Bourel \textit{et al.}~\cite{bourel2001recognition} present the first FER system under the presence of occlusion by recovering geometric facial points. 
Kotsia \textit{et al.}~\cite{kotsia2008analysis} present a comprehensive analysis on occluded FER based on Gabor features and human observers, and find that an occluded mouth degrades FER more than occluded eyes on JAFFE~\cite{lyons1998japanese} and CK~\cite{kanade2000comprehensive}. Sparse representation classifier (SRC) is widely used for artificially-occluded FER in 2010s~\cite{cotter2010sparse,zhang2012robust,cotter2010weighted}. Subsequently, a number of works handle FER with sub-region based features and fusion schemes~\cite{happy2015automatic,zhang2014facial,zhang2014random}, which detect the occlusion regions first and then remove their local features. With the popularity of data-driven deep learning techniques, several recent efforts on FER have been made on the collection of large-scale datasets~\cite{fabian2016emotionet,mollahosseini2017affectnet,li2017reliable}, and many works~\cite{Tang2013Deep,Kahou2013Combining,Liu2015AU,batista2017aumpnet,Albanie18,shanli18} \pxj{exploit deep convolutional neural networks (CNN) to improve the performance of FER.}

\pxj{We argue that explicitly removing occlusion regions is not practical since real-world occlusion is difficult to detect in itself. Directly using CNN on whole face images ignores the characteristics of occlusion and variant pose. In practices, occlusion and pose variations can lead to unseen regions of input faces, which bring difficulties for face alignment and harm the feature extraction process. Contrasted with these difficulties, human have the remarkable ability to understand facial expressions under challenging conditions. Psychological studies indicated that human can effectively exploit both local regions and holistic faces to perceive the semantics delivered through incomplete faces~\cite{yovel}. Inspired by these facts, this paper proposes a \peng{region} based deep attention architecture for pose and occlusion robust FER, which adaptively integrates visual clues from regions and whole faces. \rpxj{Specifically, we addresses the real-world pose and occlusion robust FER problem in the following aspects.}}

First, to investigate the occlusion and pose variant FER problem, \rpxj{we build six real-world test datasets from FERPlus and AffectNet, namely Occlusion-FERPlus,  Pose-FERPlus, Occlusion-AffectNet, and Pose-AffectNet, Occlusion-RAF-DB, and Pose-RAF-DB.} The occlusion test datasets are manually annotated with occlusion types of wearing mask/glasses, objects in left/right, objects in upper face, objects in bottom face. The pose-variant test datasets are automatically labeled by a recent head pose estimation toolbox~\cite{amos2016openface}. We observe that the performance of existing CNN methods degrade significantly in occlusion and pose-variant environments.


Second, we propose the Region Attention Network (RAN), to capture the importance of facial regions for occlusion and pose robust FER. \peng{The RAN is comprised of a feature extraction module, a self-attention module, and a relation attention module. The later two modules aim to learn coarse attention weights and refine them with global context, respectively.} Given a number of facial regions, our RAN \peng{learns attention weights for each regions in an end-to-end manner, and} aggregates their CNN-based features into a compact fixed-length representation. \pxj{Besides, the RAN model has two auxiliary effects on the face images. On one hand, \peng{\textit{cropping regions can enlarge} the training data which is important for those insufficient challenging samples}. On the other hand, \textit{rescaling the regions to the size of original images highlights fine-grain facial features}.} Extensive experiments indicate that our RAN significantly improves the performance of FER in occlusion and pose variant conditions.

Third, since facial expressions are mainly defined by \peng{multiple facial action units}~\cite{boucher1975facial}, we propose a Region Biased Loss (RB-Loss) to encourage a high attention weight for the most important region. Our RB-Loss resorts a simple constraint on the RAN that the maximum attention weight of facial regions should be larger than the one of the original face image. Experiments show that the RB-Loss further improves FER slightly without additional computation cost.
Our FER solution achieves state-of-the-art results on FERPlus, AffectNet, RAF-DB, and SFEW with accuracies of \textbf{89.16\%}, \textbf{59.5\%}, \textbf{86.9\%}, and \textbf{56.4\%}, respectively.


\section{Related work}
\label{related}
\pxj{In this section, we mainly present related works on normal FER problem, the occlusion and pose variant FER problem, and attention mechanism.}

\textbf{Facial Expression Recognition}.
\pxj{Generally, a FER system mainly consists of three stages, namely face detection, feature extraction, and expression recognition. 
In face detection, several face detectors like MTCNN~\cite{7553523} and Dlib~\cite{amos2016openface}) are used to locate faces in complex scenes. The detected faces can be further aligned alternatively. For feature extraction, various methods are designed to capture facial geometry and appearance features caused by facial expressions. According to the feature type, they can be grouped into engineered features and learning-based features. For the engineered features, they can be further divided into texture-based local features, geometry-based global features, and hybrid features. The texture-based features mainly include SIFT~\cite{sift}, HOG~\cite{1467360}, Histograms of LBP~\cite{SHAN2009803}, Gabor wavelet coefficients~\cite{999679}, etc. The geometry-based features are mainly based on the landmark points around noses, eyes, and mouths. Combining two or more of the engineered features refers to the hybrid feature extraction, which can further enrich the representation. 
For the learned features, Fasel~\cite{Fasel2002Robust} finds that a shallow CNN is robust to face poses and scales. Tang~\cite{Tang2013Deep} and Kahou \textit{et al}.~\cite{Kahou2013Combining} utilize deep CNNs for feature extraction, and win the FER2013 and Emotiw2013 challenge, respectively. Liu \textit{et al}.~\cite{Liu2015AU} propose a Facial Action Units based CNN architecture for expression recognition. After feature extraction, the next stage is to feed the features into a supervised classifier such as Support Vector Machines (SVMs), softmax layer, and logistic regression to assign expression categories.} 

\rpxj{To avoid overfitting on small facial expression datasets, many recent studies~\cite{meng2019frame, albanie2018emotion, tan2017group, levi2015emotion, zhao2016peak, 7961731} utilize face recognition datasets to pre-train a network, and then fine-tune it on target expression datasets. Levi and Hassner~\cite{levi2015emotion} leverage the CASIA-WebFace\cite{yi2014learning} face recognition dataset to pretrain four different VGGNet\cite{simonyan2014very} and GoogleNet\cite{szegedy2015going}. Zhao \textit{et al}.~\cite{zhao2016peak} propose a Peak Gradient Suppression (PGS) scheme for training and also pretrain their models on CASIA-WebFace. Ding \textit{et al}.~\cite{7961731} propose a FaceNet2ExpNet framework which jointly trains FER task and face recognition task. Albanie \textit{et al}. use the VGGFace model (face recognition model) and fine-tune it on FERPlus with soft probabilities. Meng \textit{et al}. evaluate different face recognition model architectures and used face recognition datasets for facial expression. }


\textbf{FER in Occlusion and Pose Variant Condition}.
Occlusion and variant pose usually occur in real-world scenarios as facial regions can be easily occluded by sunglasses, a hat, a scarf, etc.
Partial occlusion can be divided into two types according to whether the real object causes occlusion: one is artificial occlusion, and the other is real-life occlusion. Few attempts have been made on the real-world occlusion FER problem. Kotsia \textit{et al.}~\cite{kotsia2008analysis} demonstrate how artificial partial occlusion affects the FER, and discuss how to deal with it. Liu et al.~\cite{6846619} propose a novel FER method to address partial occlusion problem based on Gabor multi-orientation features fusion and local Gabor binary pattern histogram sequence (LGBPHS). Cotter \textit{et al}.~\cite{cotter2010sparse,cotter2010weighted} propose to use sparse representation classifier for partial occlusion FER. 
\peng{The latest related work~\cite{8576656} designs a patch-based attention network for occlusion aware FER. The patches are cropped from the area of eyes, nose, mouth and so on. The selected 24 patches are fed into an attention network which is near to the self-attention module in our work. Our work differs from \cite{8576656} in that i) we crop relative large regions instead of small fixed parts by considering that the facial expression is connected to multiple AUs, and ii) we refine the attention weights with a relation-attention module and region bias loss function.}
As for the pose variant FER problem, Rudovic et al.~\cite{6341749} propose the Coupled Scaled Gaussian Process Regression (CSGPR) model for head-pose normalization. 
\peng{Different from existing methods, we address both occlusion and pose variant FER problems in an end-to-end manner with an elaborately-designed region attention network architecture and collected test datasets.}

\begin{figure*}[!t]
\centering
\includegraphics[width=0.8\textwidth]{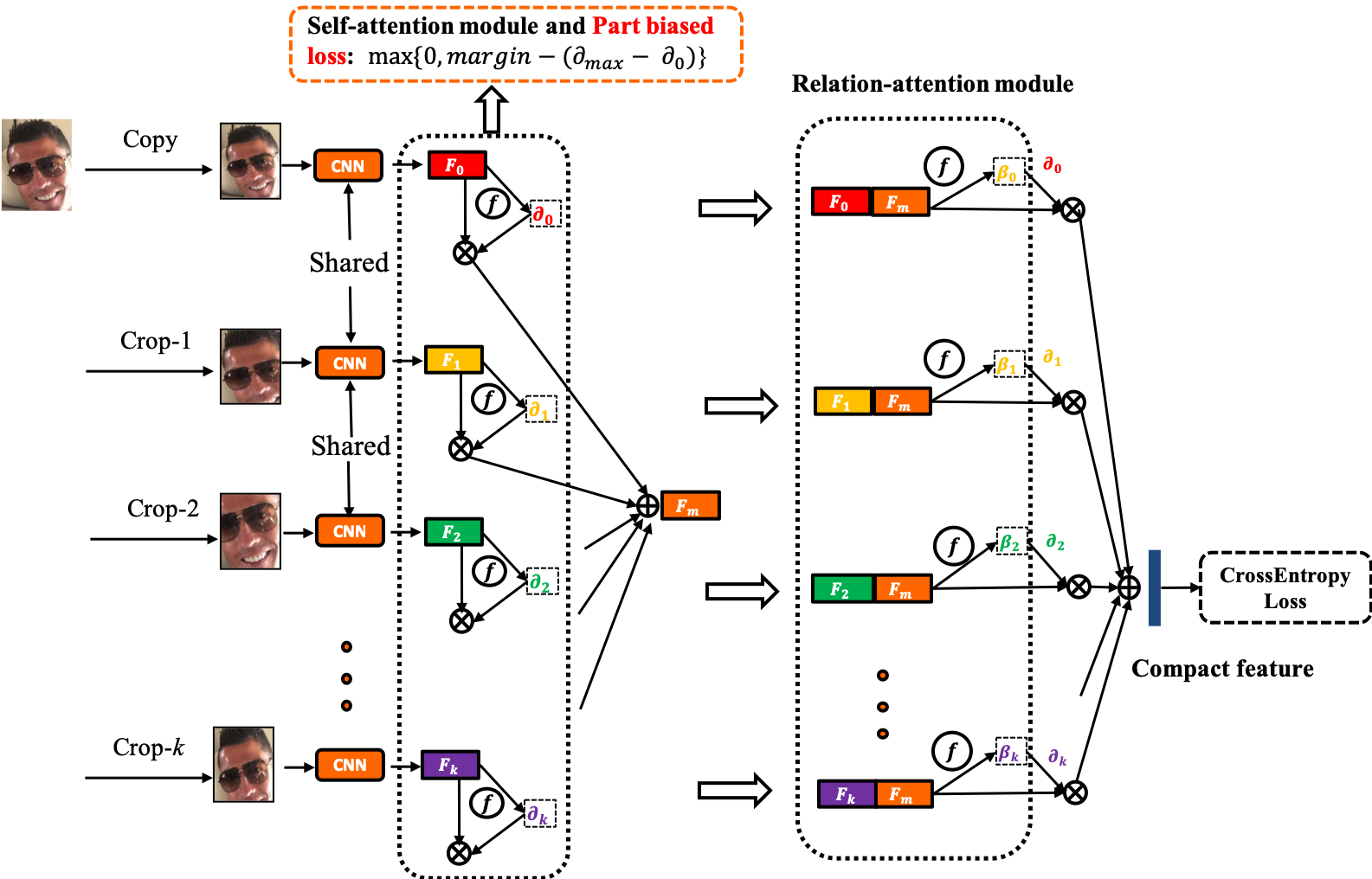}
\DeclareGraphicsExtensions.
\caption{The framework of our RAN. A face image is cropped into several regions, and these regions are fed into a backbone CNN for feature extraction. The self-attention and relation-attention module are then used to obtain compact face representation. \raisebox{.5pt}{\textcircled{\raisebox{-.9pt} {$f$}}} denotes the sigmoid function.  }
\label{fig:framework}
\end{figure*}

\textbf{Attention Networks}.
\pxj{
Attention mechanisms are firstly developed on the basis of reinforcement algorithm. Mnih et al.~\cite{NIPS2014_5542} use the attention on the RNN model for image classification, and then it is successfully utilized for machine translation tasks. Bahdanau et al.~\cite{bengio} use an attention-like mechanism to simultaneously translate and align the source languages, and their work is the first attempt to apply attention mechanism to machine translation. Afterwards, many self-attention models are proposed for different tasks, such as LSTM for machine reading~\cite{ChengDL16}, multi-head attention for machine translation~\cite{NIPS2017_7181} and attention clusters for video classification~\cite{longxiang}. 
Yu et al.~\cite{attention_wang} propose an attention network for face detection, which highlights the face regions in anchor generation step.
Perhaps the most similar work to ours is the Neural Aggregation Network (NAN) proposed by Yang et al.~\cite{YangRCWLH16}. NAN uses a cascade attention mechanism to aggregate face features of a video or set into a compact video representation. Our work differs from NAN by that self-attention and relation-attention module is used in RAN to aggregate facial region features for FER in static images, and a region biased loss is introduced to enhance region weights. }


\section{Methodology}
\label{method}
In this section, we first give an overview of our proposed region attention networks (RAN), and then detail each module and the region biased loss in RAN. We then present the region generation strategies and finally describe the collected occlusion and pose variant FER dataset.
\subsection{Overview}
As mentioned in Sec.~\ref{intro}, several early works try to detect the occlusion regions and then remove the region features to address the facial expression recognition with regional occlusion. Along with this idea, we aim to automatically reduce or eliminate the effect of occlusion and irrelated regions with an end-to-end deep architecture.

Considering both large pose and occlusion issues in facial expression recognition, we propose a Region Attention Network (RAN) to alleviate the degradation of naive face based CNN models. The proposed RAN can adaptively capture the importance of facial region information, and make a reasonable trade-off between region and global features. The pipeline of our RAN is illustrated in Figure \ref{fig:framework}. It mainly consists of three modules, namely region cropping and feature extraction module, self-attention module, and relation-attention module. Given a face image (after face detection), we first crop it into a number of regions with fixed position cropping or random cropping. We will compare these strategies in experiments. These regions along with the original face region are then fed into a backbone CNN model for region feature extraction. Subsequently, the self-attention module assigns an attention weight for each region using a fully-connected (FC) layer and the sigmoid function. An alternative region biased loss (RB-Loss) is further introduced to regularize the attention weights and enhance the most valuable region in self-attention module. We aggregate these region features to a global representation ($F_m$ in Figure \ref{fig:framework}). Then the relation-attention module uses a similar attention mechanism on the concatenation of individual region feature and global representation to further capture content-aware attention weights.
Finally, we leverage the weighted region feature and the global representation to predict the expressions.


\begin{figure*}[!t]
\centering
\includegraphics[width=0.8\textwidth]{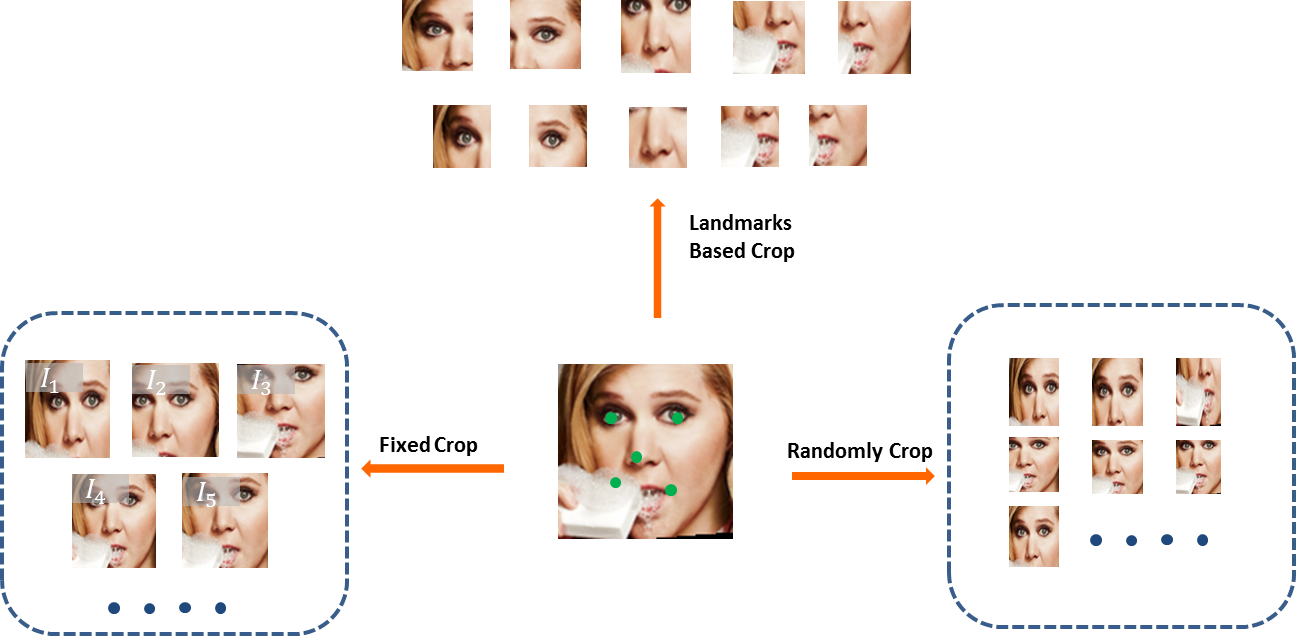}
\caption{An example of our region generation methods. Left: fixed position cropping. Right: random cropping. Upper: landmark-based cropping.}
\label{fig:get_region}
\end{figure*}


\subsection{Region Attention Networks}
As shown in Figure \ref{fig:framework}, the proposed RAN mainly consists of two stages. The first stage is to coarsely calculate the importance of each region by a FC layer conducted on its own feature, which is called self-attention module. The second stage seeks to find more accurate attention weights by modeling the relation between the region features and the aggregated content representation from the first stage, which is called relation-attention module. 

Formally, we denote a face image as $I$, its duplicate as $I_0$, and its crops as ${I_1},{I_2}, \cdots ,{I_k}$,  and the backbone CNN as $r(\cdot;{\theta})$. The feature set $X$ of $I$ is defined by:
\begin{multline}
    X = [{F_0},{F_1}, \cdots, {F_k}] = [r({I_0};{\theta}),r({I_1};{\theta}), \cdots ,r({I_k};{\theta})],
\end{multline}
where $\theta$ is the parameter of backbone CNN. 

\textbf{Self-attention module}. 
With these region features, the self-attention module applies a FC layer and a sigmoid function to estimate coarse attention weights. Mathematically, the attention weight of the $i$-th region is defined by: 
\begin{equation}
\mu_{i} = f(F_i^\top\mathbf{q}^0),
\end{equation}
where $\mathbf{q}^0$ is the parameter of FC, $f$ denotes the sigmoid function. 
In this stage, we summarize all the region features with their attention weights into a global representation $F_m$ as follows,
\begin{equation}
\label{eq:fm}
\kwang{F_{m} = \frac {1}{\sum_{i=0}^{n}\mu_{i}}\sum_{i=0}^{n}\mu_{i}F_i.}
\end{equation}
$F_m$ is a compact representation and can be used as the final input of classifier. We compare the self-attention aggregation to the straightforward average pooling and concatenation (with fixed number of crops) in Sec. \ref{experiment}.

\textbf{Relation-attention module}.
The self-attention module learns weights with individual features and non-linear mapping, which is rather coarse. Since the aggregated representation $F_m$ inherently represents the contents of all the facial regions, the attention weights can be further refined by modeling the relation between region features and this global representation $F_m$.

Inspired by the global attention in neural machine translation~\cite{luong2015effective} and the relation-Net in low-shot learning~\cite{yang2018learning}, we use the sample concatenation and another FC layer to estimate new attention weights for region features. The new attention weight of the $i$-th region in relation-attention module is formulated as,
\begin{equation}
\nu_{i} = f([F_i:F_m]^\top\mathbf{q}^1),
\end{equation}
where $\mathbf{q}^1$ is the parameter of FC, and $f$ denotes the sigmoid function. In this stage, we aggregate all the region information along with the coarse global representation from self-attention into a new compact feature as,
\begin{equation}
\kwang{P_{RAN} = \frac {1}{\sum_{i=0}^{n}\mu_{i}\nu_{i}}\sum_{i=0}^{n}\mu_{i}\nu_{i}[F_{i}:F_m].}
\end{equation}
$P_{RAN}$ is used as the final representation of the proposed RAN method. 

\textbf{Region Biased Loss}.
Inspired by the observation that different facial expressions are mainly defined by different facial regions~\cite{boucher1975facial}, we make a straightforward constraint on the attention weights of self-attention, \textit{i.e.} region biased loss (RB-Loss).
This constraint enforces that one of the attention weights from facial crops should be larger than the original face image with a margin. For example, the Crop-2 in Figure \ref{fig:framework} can be more discriminative than the original one.
Formally, the RB-Loss is defined as,
\begin{equation}
\mathcal{L}_{RB} = \max\{0, \alpha - (\mu_{max} - \mu_{0})\},
\end{equation}
where $\alpha$ is a hyper-parameter served as a margin, $\mu_{0}$ is the attention weight of the copy face image, $\mu_{max}$ denotes the maximum weight of all facial crops.

In training, the classification loss is jointly optimized with the region biased loss. The proposed RB-Loss enhances the effect of region attention and encourages RAN to obtain superior weights of region and global representations. In fact, the RB-Loss can be also added to the relation-attention module. However, since the features of the relation-attention module already include holistic information, we experimentally find there is no gain by adding RB-Loss on the relation attention module.

\begin{figure*}[htp]
\centering
\includegraphics[width=.98\linewidth]{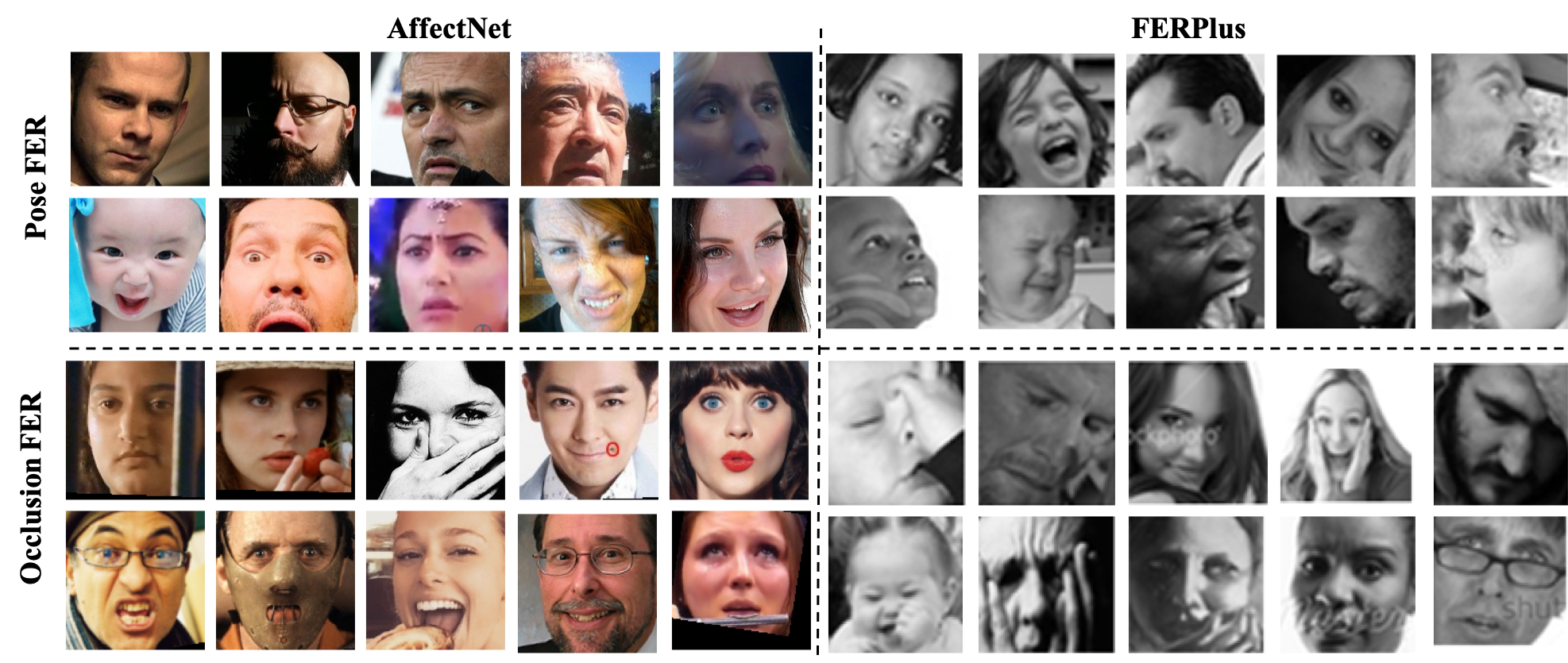}  
\caption{Some examples of our collected occlusion and pose variant test datasets. The left color images are from the test set of AffectNet, and the right gray images are from  the test set of  FERPlus.}
\label{fig:pose_occlusion_dataset}
\end{figure*}

\subsection{Region Generation}
Cropping multiple regions is a fundamental step of our RAN. Too large regions lead to the reduced diversity of features and degrade to the case of many duplicates of the original face. Too small regions lead to insufficient discrimination ability of region features. In this paper, we evaluate three kinds of region generation schemes for our region attention networks, namely fixed position cropping,  random cropping, and landmark-based cropping which are depicted in Figure \ref{fig:get_region}. 

\textbf{Fixed position cropping}.
Since the face image can be well aligned by the recent advanced face alignment methods, a simple region generation scheme is to crop regions in fixed positions with fixed scales. Specifically, \pxj{we crop five regions. Three of them are the top-left, top-right and center-down face regions, which have fixed size of 0.75 scale ratio of the original face. The other two regions are similar to those used in the smile-classification task of \cite{zhang2016gender}. Here we crop the center regions with sizes of 0.9, and 0.85 scale ratio of the original face.} All the crops are resized to have the same input size of the backbone CNN. 

\textbf{Random cropping}.
In deep face recognition, the DeepID method uses 200 random crops for each face image to enhance its performance~\cite{sun2014deep}. For random cropping in our approach, we randomly crop $N$ regions with random sizes ranged from 0.7 to 0.95 scale ratio of the original face.

\textbf{Landmark-based cropping}. Given facial landmarks, a straightforward method  is to crop regions surrounding them, which is also used in \cite{8576656}.  Here we use MTCNN to detect five facial landmarks (i.e. left eye, right eye, nose, left mouth corner, and right mouth corner), and use them to crop five regions. Specifically, according to each facial landmark, we use a radius $r$ to crop regions and remove the regions which are out of the original image. 


\subsection{Occlusion and Pose Variant Dataset}
Though our proposed RAN can be used for FER in any conditions, we focus on the real-world occlusion and pose variantion problems. \peng{ To the best of our knowledge, there is only a small real-world occlusion test dataset released in \cite{8576656} very recently,} and there is no publicly available facial expression dataset that addresses both occlusion and pose annotations. To examine our method under real-world
scenario, \rpxj{we build six test datasets from the existing large-scale FER datasets.
From the test set of FERPlus~\cite{barsoum2016training}, the validation set of AffectNet~\cite{mollahosseini2017affectnet}, and the test set of RAF-DB~\cite{8453893}, we collect the Occlusion-FERPlus, Pose-FERPlus, Occlusion-AffectNet, Pose-AffectNet, Occlusion-RAF-DB, and Pose-RAF-DB for testing.} The test set will be available at \href{https://github.com/kaiwang960112/Challenge-condition-FER-dataset}{https://github.com/kaiwang960112/Challenge-condition-FER-dataset}.  These real-world test sets are annotated with different occlusion types and different pose degrees. Some examples are illustrated in Figure \ref{fig:pose_occlusion_dataset}.

For the pose variant test sets, we use the popular OpenFace toolbox~\cite{amos2016openface} to estimate the Euler Angle in pitch, yaw, roll directions. Since the roll angle is in-plane which can be eliminated by face alignment, we only consider the pose in pitch and yaw directions. \kwang{Those faces with pitch or yaw angle larger than $30^o$ are collected to Pose-FERPlus, Pose-AffectNet, and Pose-RAF-DB.}

For the occlusion test sets, we first define several occlusion types, namely wearing mask, wearing glasses, objects in left/right, objects in upper face and objects in bottom face, non-occlusion. \kwang{Then we manually assign these categories to the test sets of FERPlus, AffectNet, and RAF-DB.} Images with at least one type of occlusion are selected as the occlusion test sets.

\kwang{We present the statistics of our collected test sets in Table \ref{tab:stat}. Among all the occlusion types on FERPluse, AffectNet, and RAF-DB, the upper occlusion has the smallest samples. The total numbers of occlusion samples in FERPlus (test) , AffectNet (validation), and RAF-DB (test) are respectively 605, 682 , and 735, which are 16.86\%, 17.05\%, and 23.9\% of their original sets. For the variant pose issue, about one-third of FERPlus (test), about two-fifths of RAF-DB, and about half of AffectNet (validation) have poses larger than 30 degrees (in pitch or yaw).}


\begin{table}[!t]
\centering
\caption{\textcolor[rgb]{0.00,0.00,0.00}{Statistics of collected test datasets.}}
\scriptsize
\begin{tabular}{@{}ccccccc@{}}
\hline
\toprule
                     & \multicolumn{4}{c}{Occlusion}                                                             & \multicolumn{2}{c}{Pose(pitch/yaw)}                                           \\ \midrule
                     & upper                & bottom               & left/right           & glasses/mask         & \textgreater{}30     & \textgreater{}45       \\
FERPlus              &          70            &         138             &                   213   &            184          &       1171    &   634       \\
AffectNet            &        84 &            183 &           128           &    288                  &          1949            &          985            \\
RAF-DB            &        126 &            151 &           160           &    298                  &          1248            &          558            \\
\bottomrule
\hline
\end{tabular}
\label{tab:stat}
\end{table}

\section{Experiments}
\label{experiment}
In this section, we first describe the used datasets and our implementation details. We then present our collected occlusion and pose variant test datasets and evaluate our proposed RAN on them. We further explore each components of RAN on FERPlus~\cite{barsoum2016training}, AffectNet~\cite{mollahosseini2017affectnet} , and SFEW~\cite{6130508}. Finally, we compare our method to the state-of-the-art approaches.

\subsection{Datasets}
\kwang{To evaluate our method, we use four popular in-the-wild facial expression datasets, namely FERPlus~\cite{barsoum2016training}, AffectNet~\cite{mollahosseini2017affectnet}, RAF-DB~\cite{8453893}, and SFEW~\cite{6130508}. These datasets cover different scales of face images and the challenging conditions. Besides, we also build occlusion and pose variant test datasets from FERPlus, AffectNet, and RAF-DB.}

\textbf{FERPlus}~\cite{barsoum2016training}. The FERPlus is extent from FER2013~\cite{goodfellow2013challenges} introduced during the ICML 2013 Challenges in Representation Learning. It is a large-scale and real-world datasets collected by the Google search engine, and consists of 28,709 training images, 3,589 validation images and 3,589 test images. All face images in the dataset are aligned and resized to 48$\times$48. The main difference between FER2013 and FERPlus is the annotation. FER2013 is annotated with seven expression labels (neutral, happiness, surprise, sadness, anger, disgust, fear) by one tagger, while FERPlus adds a \textit{contempt} label and is annotated by 10 labels. In \cite{barsoum2016training}, the authors evaluate several training schemes, such as one-hot label (majority voting) and  label distribution with cross-entropy loss. We mainly report the overall accuracy on the test set with supervision of majority voting and label distribution.

\textbf{AffectNet}~\cite{mollahosseini2017affectnet}.  The AffectNet is by far the largest dataset that provides both categorical  and Valence-Arousal annotations. The dataset contains more than one million images from Internet by querying expression-related keywords in three search engines, of which 450,000 images are manually annotated with eight basic expression labels as FERPlus. AffectNet has an imbalanced test set, a balanced validation set, and an imbalanced training set. We mainly report accuracy on the validation set where each category contains 500 samples.

\textbf{SFEW}~\cite{6130508}. The Static Facial Expressions in the Wild (SFEW) dataset is built by selecting frames from AFEW~\cite{dhall2012collecting}, which covers unconstrained facial expressions, varied head poses, large age range, occlusions, varied focus, different resolution of the face and real-world illumination. {We use the newest version of SFEW in \cite{dhall2015video} where it has been divided into three sets: train (958 images), validation (436 images), and test (372 images).} Each image is labeled with one of the seven expressions including angry, disgust, fear, happy, sad, surprise, and neutral by two independent labelers. We mainly report our performance on the validation set.

\peng{\textbf{RAF-DB}. RAF-DB~\cite{8453893} contains 30,000 facial images annotated with basic or compound expressions by 40 trained human coders. In our experiment, only images with basic emotions were used, including 12,271 images as training data and 3,068 images as test data.}

\subsection{Implementation details}
{In all the following experiments, we use the CNN detector and the ERT~\cite{kazemi2014one} based face alignment method in Dlib toolbox\footnote{http://dlib.net} to crop and align faces, and then resize them to the size of 224$\times$224.
We implement our methods with Pytorch toolbox\footnote{https://pytorch.org/}.
For the backbone CNN, we mainly use the ResNet-18~\cite{he2016deep} and VGG16~\cite{Parkhi15}. The ResNet-18 is pre-trained on MS-Celeb-1M face recognition dataset and VGG16 is downloaded from website\footnote{\url{http://www.robots.ox.ac.uk/~vgg/software/vgg$_$face/}}.
The last pooling layer of ResNet-18, and the first FC feature of VGG16 is used for facial representation.
In training phase of fixed cropping, we use all the five regions along with original face for each face image (i.e. $k=5$ in Figure \ref{fig:framework}). For training with random cropping, we replace the fixed five regions with randomly cropped ones.
When jointly training with RB-Loss and Cross-Entropy loss, the default loss weight ratio is 1:1. On all datasets, the learning rate is initialized as 0.01, and divided by 10 after 15 epochs and 30 epochs. We stop training in 40 epochs.
The margin in RB-Loss is default as 0.02.}


\begin{table}[t]
\center
\caption{\textcolor[rgb]{0.00,0.00,0.00}{Performance comparison between the proposed RAN and baseline method with occlusion and variant pose conditions.}}
\begin{tabular}{@{}cccccc@{}}
\hline
\toprule
FERPlus         & Occlusion & Pose(30) & Pose(45) \\ \midrule
Baseline &     73.33      &        78.11       &       75.50        \\
RAN (w RB-Loss) &     \textbf{83.63}      &        \textbf{82.23}          &      \textbf{80.40 }          \\
\hline
\toprule
AffectNet       & Occlusion & Pose(30) & Pose(45) \\\midrule
Baseline        &     49.48      &       50.10         &       48.50      \\
RAN (w RB-Loss) &    \textbf{58.50}     &         \textbf{53.90}      &      \textbf{ 53.19 }      \\ 
\toprule
RAF-DB         & Occlusion & Pose(30) & Pose(45) \\ \midrule
Baseline &     80.19      &        84.04       &       83.15        \\
RAN (w RB-Loss) &     \textbf{82.72}      &        \textbf{86.74}          &      \textbf{85.20 }          \\ \bottomrule
\end{tabular}
\label{tab:opose}
\end{table}

\begin{figure*}[t]
\captionsetup[subfigure]{labelformat=empty}
\begin{subfigure}{.24\textwidth}
  \centering
  \includegraphics[width=1\linewidth]{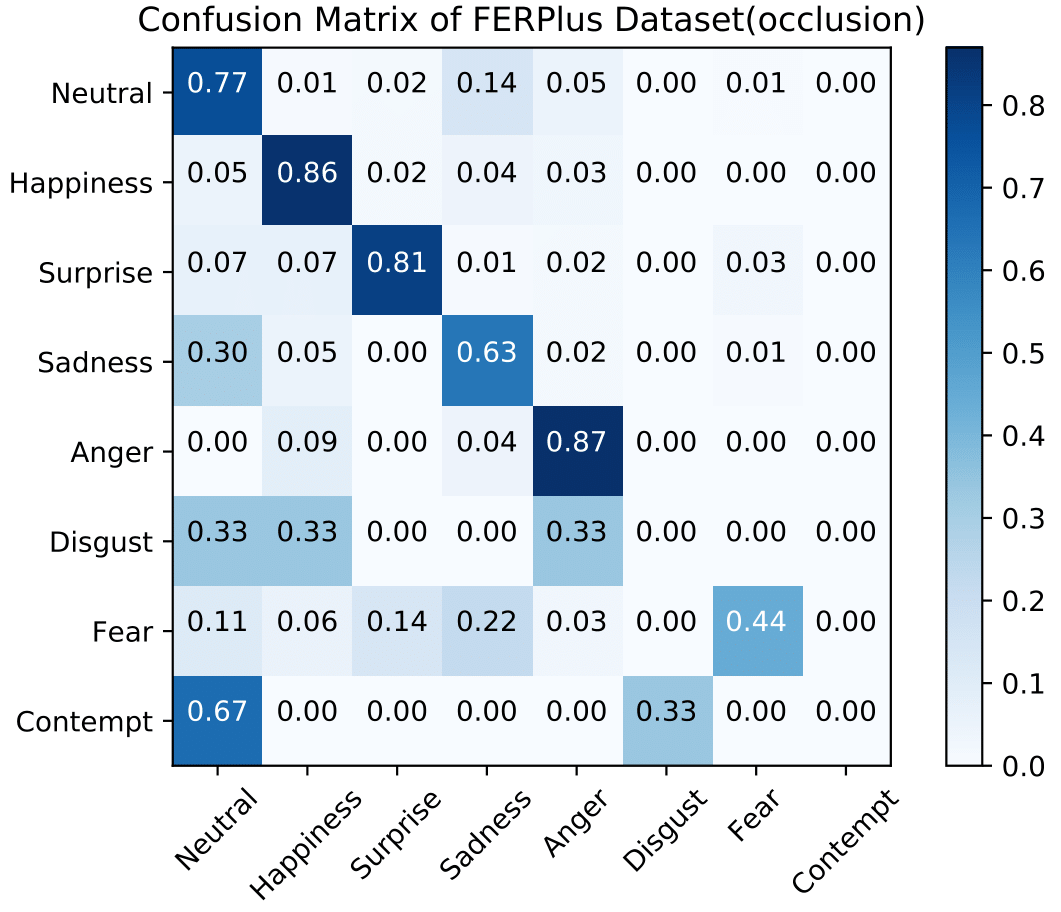}  
  \caption{\small{Baseline on Occlusion-FERPlus.}}
  \label{fig:sub-first}
\end{subfigure}
\begin{subfigure}{.24\textwidth}
  \centering
  \includegraphics[width=1\linewidth]{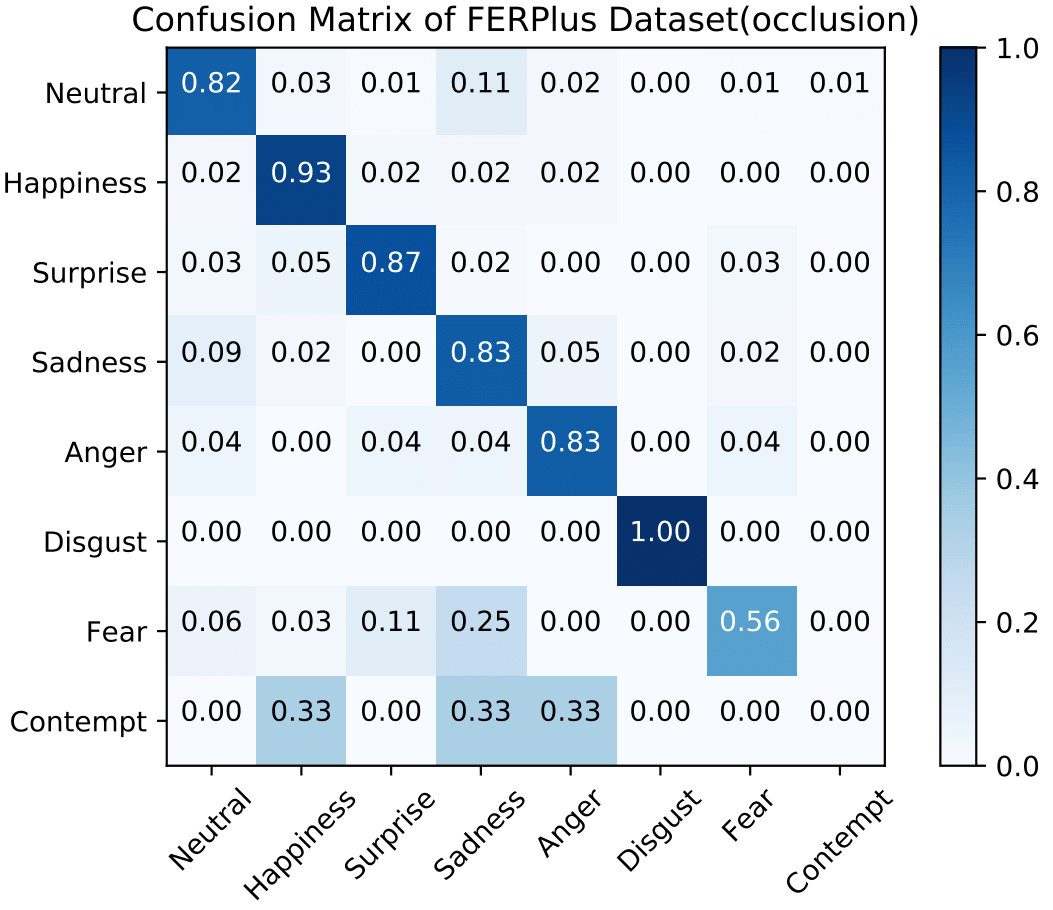}  
  \caption{\small{ RAN on Occlusion-FERPlus.}}
  \label{fig:sub-second}
\end{subfigure}
\begin{subfigure}{.24\textwidth}
  \centering
  \includegraphics[width=1\linewidth]{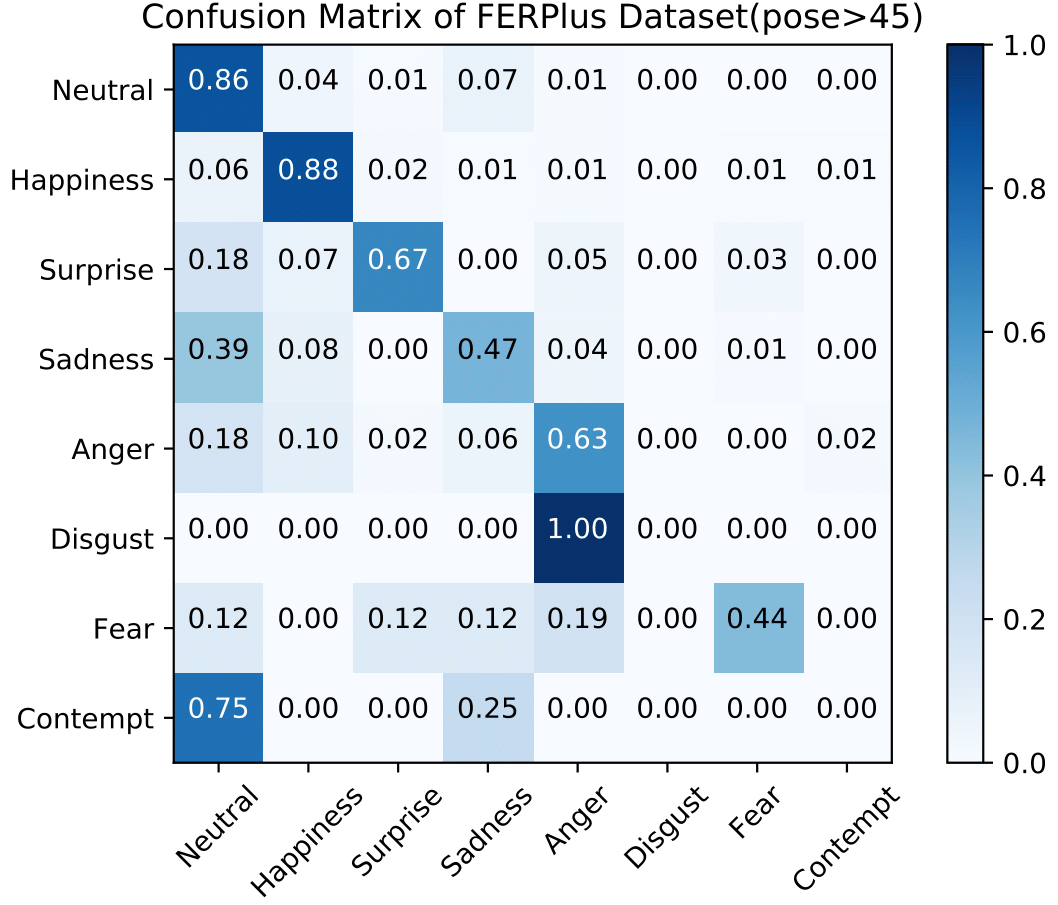}  
  \caption{\small{Baseline on Pose-FERPlus.}}
  \label{fig:sub-third}
\end{subfigure}
\begin{subfigure}{.24\textwidth}
  \centering
  \includegraphics[width=1\linewidth]{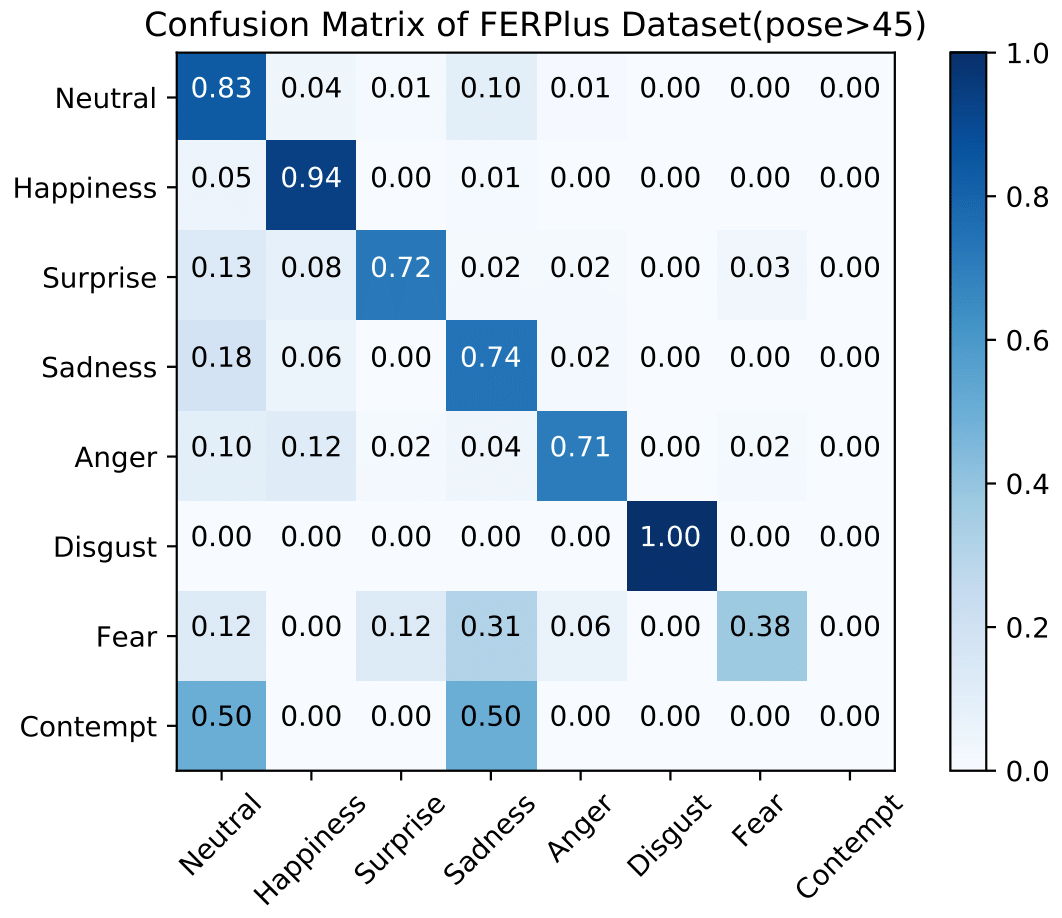}  
  \caption{\small{ RAN on Pose-FERPlus.}}
  \label{fig:sub-forth}
\end{subfigure}
\caption{The confusion  matrices of baseline methods and our RAN on the Occlusion- and Pose-FERPlus test sets.}
\label{fig:confusion_metrics_ferplus}
\end{figure*}

\begin{figure*}[t]
\captionsetup[subfigure]{labelformat=empty}
\begin{subfigure}{.24\textwidth}
  \centering
  \includegraphics[width=1\linewidth, height=0.9\linewidth]{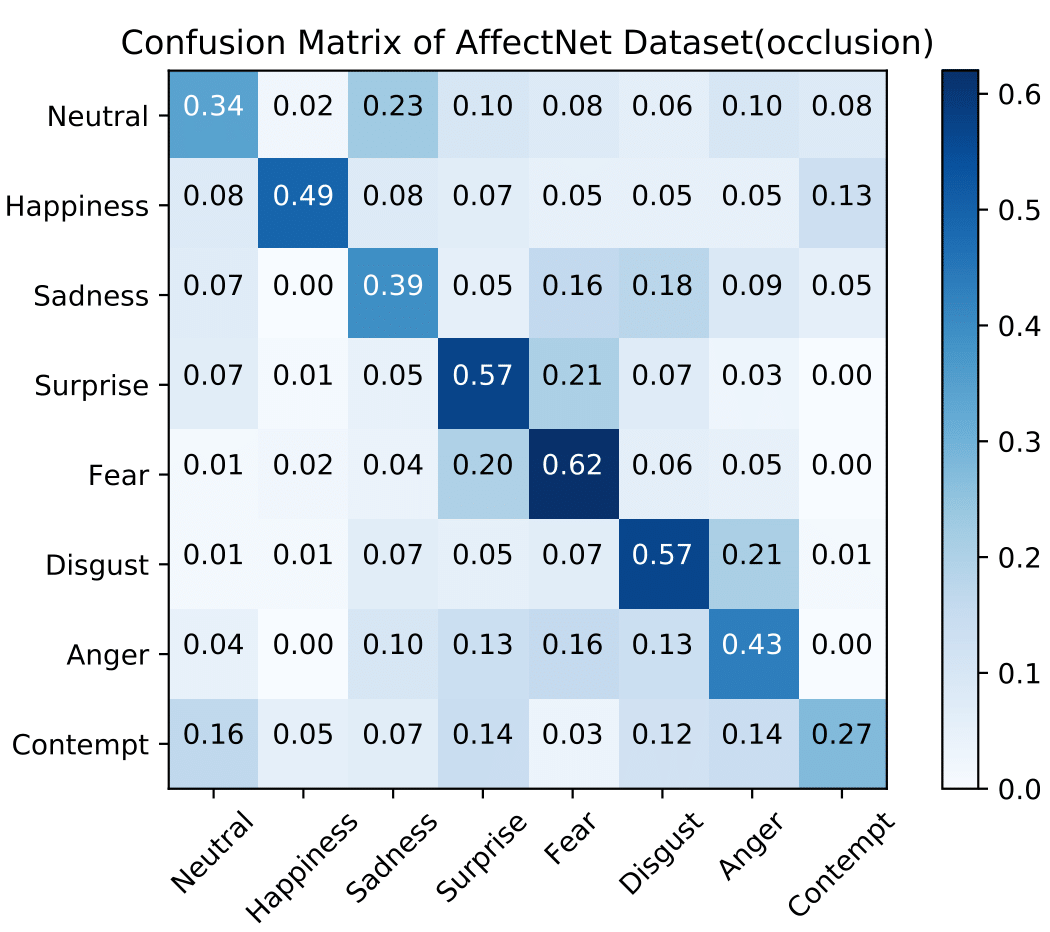}  
  \caption{\small{Baseline on Occlusion-AffectNet.}
  \label{fig:sub-first}}
\end{subfigure}
\begin{subfigure}{.24\textwidth}
  \centering
  \includegraphics[width=1\linewidth, height=0.9\linewidth]{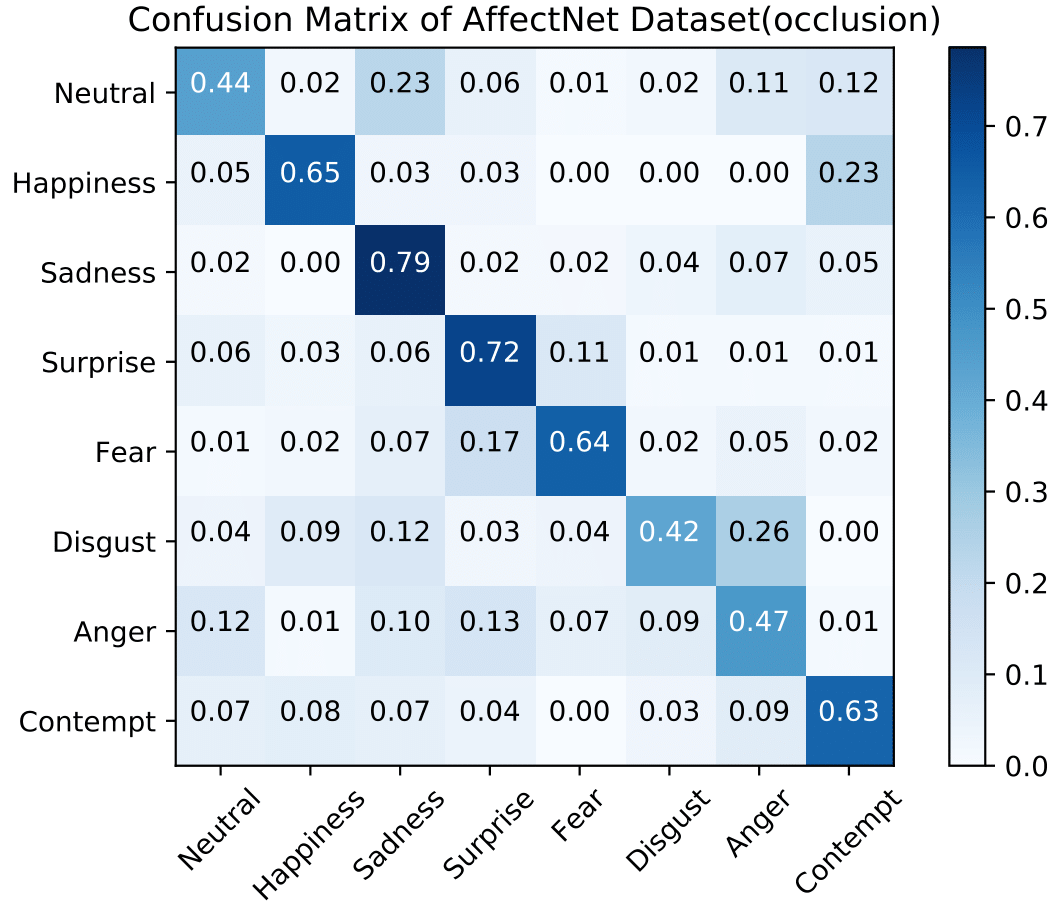}  
  \caption{\small{RAN on Occlusion-AffectNet.}}
  \label{fig:sub-second}
\end{subfigure}
\begin{subfigure}{.24\textwidth}
  \centering
  \includegraphics[width=1\linewidth, height=0.9\linewidth]{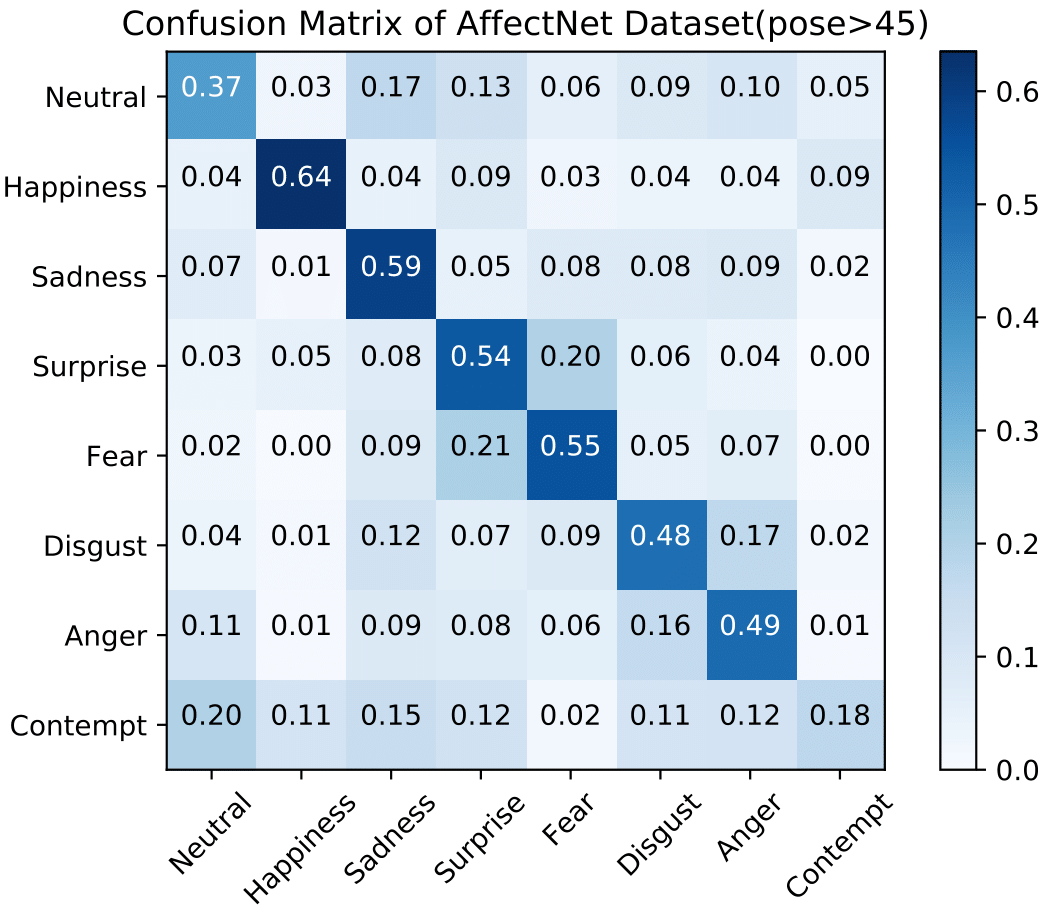}  
  \caption{\small{Baseline on Pose-AffectNet.}}
  \label{fig:sub-third}
\end{subfigure}
\begin{subfigure}{.24\textwidth}
  \centering
  \includegraphics[width=1\linewidth, height=0.9\linewidth]{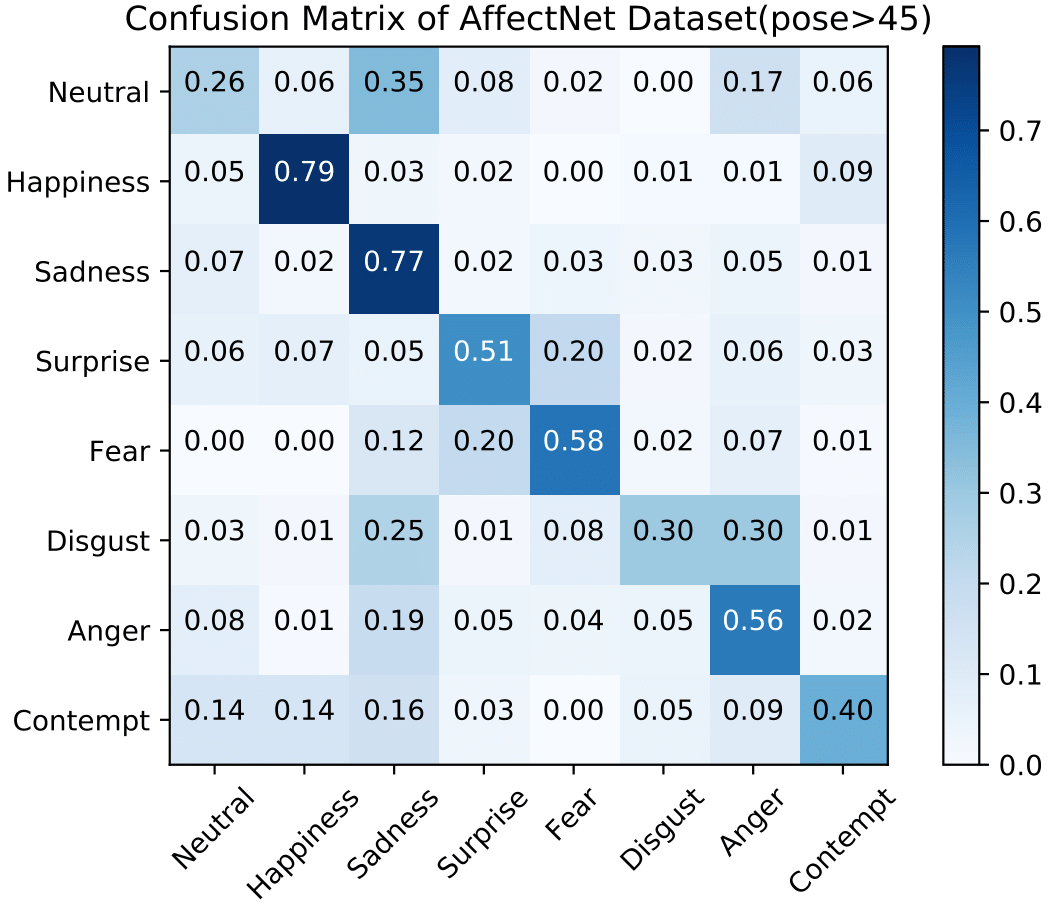}  
  \caption{\small{RAN on Pose-AffectNet.}}
  \label{fig:sub-forth}
\end{subfigure}
\caption{The confusion matrices of baseline methods and our RAN on the Occlusion- and Pose-AffectNet test sets.}
\label{fig:confusion_metrics_affectnet}
\end{figure*}

\subsection{FER with occlusion and variant pose in the wild}
To address the occlusion and pose variant issues, we construct several test subsets with occlusion and pose annotations, \textit{i.e.} Occlusion-FERPlus,  Pose-FERPlus,  Occlusion-AffectNet, Pose-AffectNet, \rpxj{Occlusion-RAF-DB, and Pose-RAF-DB.}
We evaluate our RAN on the collected datasets with the default setting (\textit{ i.e}. ResNet18 with alignment, RB-Loss and fixed cropping). We fine-tune the ResNet18 on original face images as baselines. Table \ref{tab:opose} presents the comparison between the baselines and our method. 
\kwang{Our RAN improves the baseline method significantly, with gains 10.3\%, 10.02\% and 2,53\% on Occlusion-FERPlus, Occlusion-AffectNet and Occlusion-RAF-DB, respectively. On Pose-FERPlus, Pose-AffectNet and Pose-RAF-DB, the RAN also outperforms the baseline with a large margin. Specifically, with pose larger than 30 degrees, the gains are 4.12\%, 3.09\% and 2.70\% on Pose-FERPlus, Pose-AffectNet and Pose-RAF-DB, respectively. The gains are improved to 4.9\%, 5.4\% and 2.05\% with pose larger than 45 degrees. Overall, these results demonstrate the effectiveness of our proposed RAN on occlusion and variant pose FER data.}

We present the confusion matrices of our RAN and these baselines in Figure \ref{fig:confusion_metrics_ferplus} and Figure \ref{fig:confusion_metrics_affectnet} to further investigate our improvements. 
We find that our RAN consistently boosts the ``happiness'', ``superise'', and ``sadness'' categories on all the test sets. It may be explained by that these facial expressions have clear region features, such as action units of  ``Lip Corner Puller'' , ``Cheek Raiser'', and ``Lip Corner Depressor'', which can be effectively captured by our RAN. 
%

We also conduct a fair comparison on the recent occlusion test dataset: FED-RO~\cite{8576656}. We use the RAN with default setting, and train it using the same training data as \cite{8576656}. We finally achieve \textbf{67.98}\% which is clearly better than 66.5\% of \cite{8576656}.

\begin{table}[t]
\center
\caption{\rpxj{The performance of individual regions with occlusion and variant pose conditions on FERPlus. `Aug. Training' means that we augment the dataset by combining all the regions and original images  and then train the network.}}
\begin{tabular}{@{}cccccc@{}}
\hline
\toprule
Region         & Occlusion & Pose(30) & Pose(45) \\ \midrule
Original ($I_0$)       &     73.33      &        78.11       &       75.50        \\
$I_1$        &    67.43       &        74.27       &       71.40
     \\
$I_2$          &    64.13       &        72.22       &       70.30
     \\
$I_3$         &    72.22       &        78.48       &       76.84
     \\
$I_4$          &    72.8       &        78.54       &       77.00
     \\
$I_5$         &    74.54       &       78.63       &       75.35
     \\
Score Fusion ($I_0-I_5$)  &   75.70      &         79.84      &        78.45
      \\
Aug. Training ($I_0-I_5$)  &   79.92      &         81.24      &     79.26
      \\
RAN (w RB-Loss) &     \textbf{83.63}      &        \textbf{82.23}          &       \textbf{80.40 }          \\
 \bottomrule
\end{tabular}
\label{tab:region}
\end{table}

\begin{figure*}[t]
\begin{subfigure}{.5\textwidth}
  \centering
  \includegraphics[width=1\linewidth]{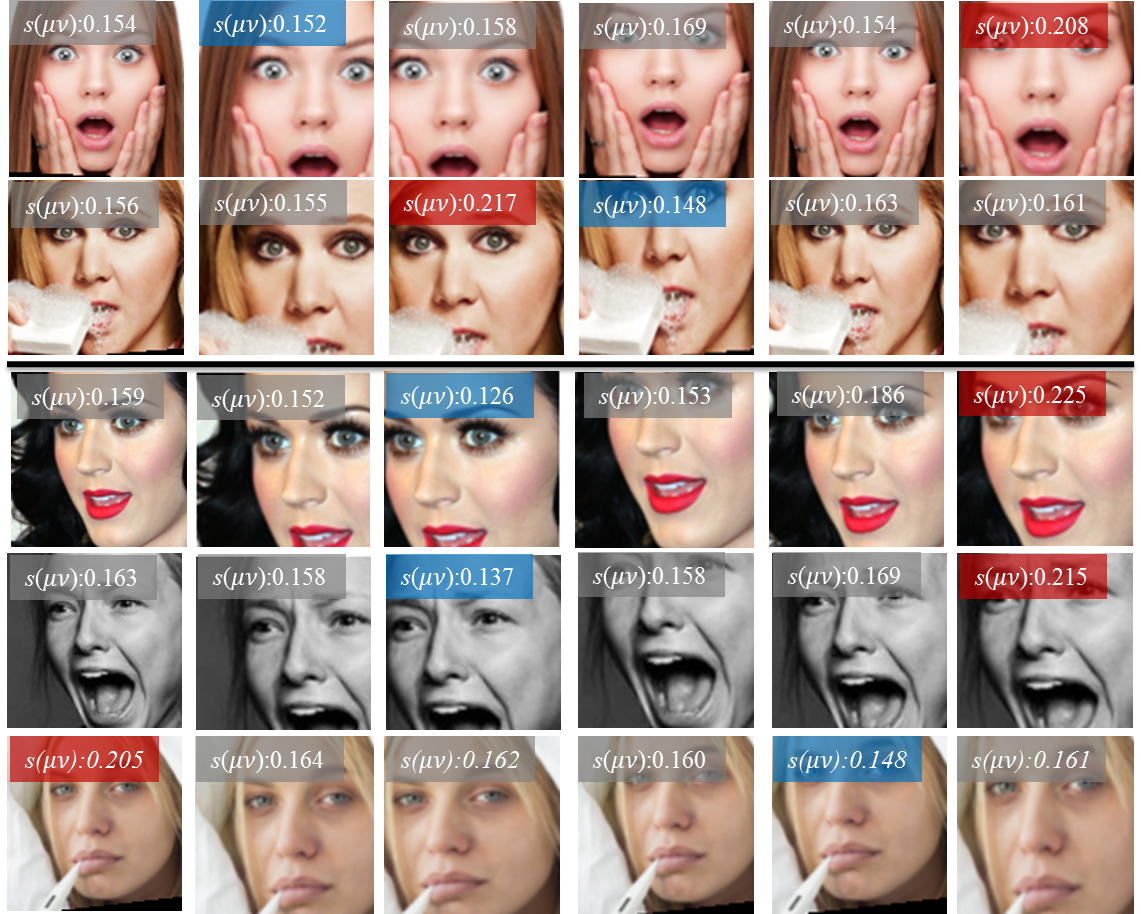}  
  \caption{RAN with RB-Loss }
  \label{fig:sub-first}
\end{subfigure}
\begin{subfigure}{.5\textwidth}
  \centering
  \includegraphics[width=1\linewidth]{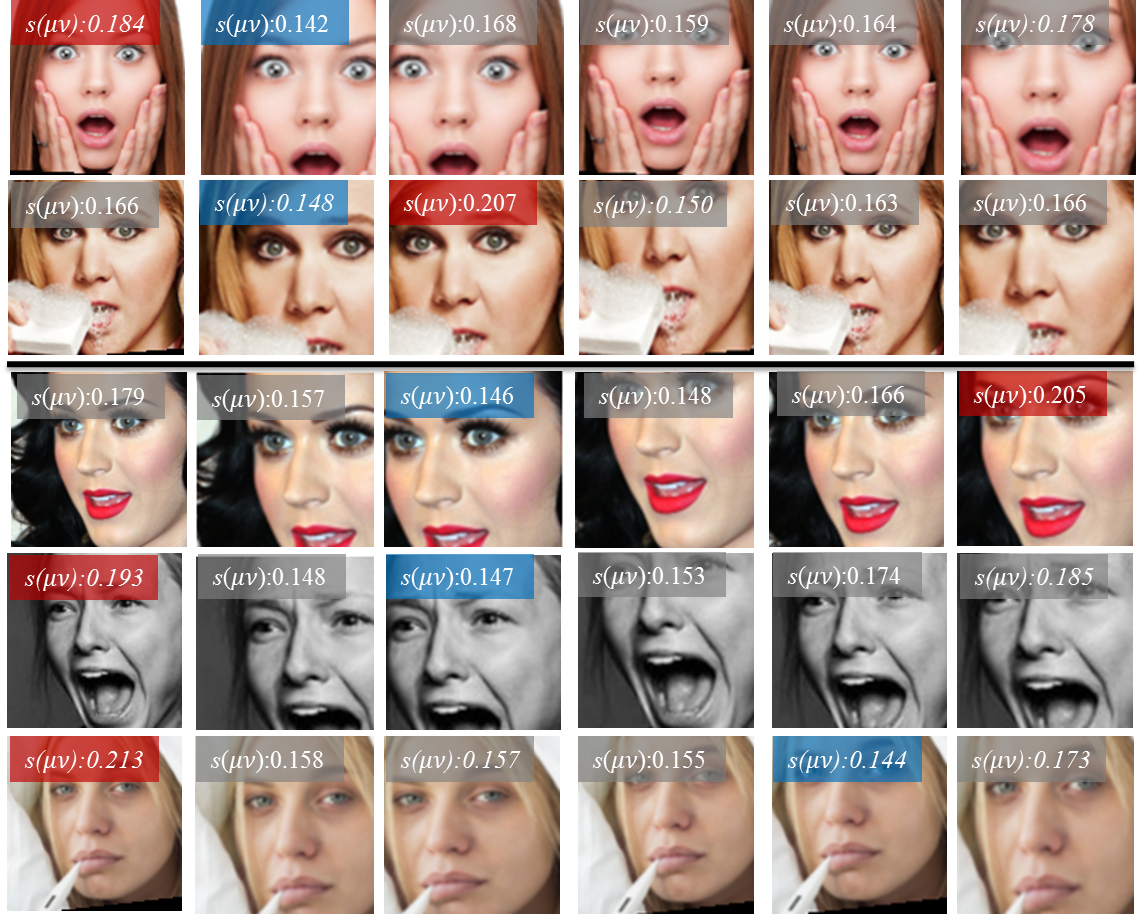}  
  \caption{RAN without RB-Loss }
  \label{fig:sub-first}
\end{subfigure}
\caption{Illustration of learned attention weights for different regions along with origianl faces. $s(\cdot)$ denotes the softmax function. Red-filled boxes indicate the highest weights while blue-filled ones are the lowest weights. From left to right, the columns represent the original faces, regions $I_1$ to $I_5$. Note that the left and right figures show the weights obtained with and without the RBLoss respectively. Better viewed in PDF.}
\label{fig:score4poseocc}
\end{figure*}

\textbf{\peng{Individual regions and their combination.}}
\rpxj{Since our RAN integrates several regions in a single network, we present the performace of individual regions and their score fusion on Occlusion- and Pose-FERPlus in Table \ref{tab:region}. To investigating if the improvement of our RAN only comes from augmented data, we also train a traditional model by mixing all the regions and the original images for data augmentation, which is called \textit{aug. training}. We conclude that i) the performance of individual regions are comparable to each other except for the region $I_1$ and $I_2$, ii) a naive score fusion (i.e. average) and mixing all the regions improve individual performance slightly, and iii) our RAN outperforms the score fusion and Aug. Training by a large margin. Compared to score fusion and fusion training, our RAN takes account of the importance of region features and also emphasises the most important region with RB-Loss.}


\textbf{What is learned for occlusion and pose variant faces}?
To better explore our RAN, we illustrate the final attention weights for several examples with RB-Loss and without RB-Loss in Figure \ref{fig:score4poseocc}(a) and Figure \ref{fig:score4poseocc}(b), respectively. 
\kwang{Occlusion examples are shown in the first two rows, and pose examples in the third and forth rows. We also show a bad example in the last row.}

For the occlusion examples, our RAN with RB-Loss gets the highest weight on the small center crop (i.e. $I_5$) for the first example. It makes sense since this image suffers from the left and right occlusion. In the second example which suffers from bottom-left occlusion, the RAN with RB-Loss automatically assigns the highest weight to the up-right region while suppresses the bottom-left region. 
\kwang{For both pose-variant examples in the third and forth rows, our RAN with RB-Loss gets high attention weights on center regions while gets low weights on the up-right regions.} This may be explained by that the up-right regions contain the most of irrelated information on the near-profile faces. With RB-Loss and RAN, the original faces get relatively average attention weights among all the examples.
\kwang{For a bad example in the last row, our RAN with RB-Loss does not assign the regions with high weights. It may illustrate that the tiny occlusion can not impact the FER system a lot. The attention weights of tiny occlusion face are more likely to be random.}
Compared to the RB-Loss case, though RAN without RB-Loss can also assign different attention weights similarly, the weights for all the regions from RAN without RB-Loss are smoother. In addition, the original image prefers to have the highest weight without RB-Loss.
%

\subsection{Ablation study on FERPlus and AffectNet} 
To validate the generality of our method, we conduct an ablation study on the full test set of FERPlus and the full validation set of AffectNet with default setting.
Face alignment is a standard pre-processing method for face analysis, while a few works do not utilize~\cite{barsoum2016training,mollahosseini2017affectnet} for FER task. Here we also  study the effect of face alignment.

\textbf{Attention modules}. We first study the attention modules of our RAN without using RB-Loss. The evaluation results on FERPlus and AffectNet are presented on Table \ref{tab:FERPlus} and Table \ref{tab:AffectNet}, respectively. On FERPlus without face alignment, the self-attention ($F_m$ in Eq. (\ref{eq:fm})) improves the baseline by 0.4\%. Adding the relation-attention module, our method outperforms the baseline by 1.13\% and 3.05\% on FERPlus and AffectNet without face alignment. Face alignment is found to significantly boost the baseline method on both datasets, while its effect is limited when using our proposed RAN. 
This can be explained by that our method implicitly learns to align facial regions with the attention mechanism as that in machine traslation~\cite{bengio}.
With face alignment, our attention modules improve the baselines by 0.83\% and 1.85\% on FERPlus and AffectNet, respectively.

\begin{table}[t]
\center
\caption{Evaluation of all components of our RAN along with face alignment on FERPlus.}
\label{tab:FERPlus}
\begin{tabular}{@{}cccccc@{}}
\toprule
Align & Self-att. & Relation-att. & RB-Loss & Accuracy  \\
\midrule
 & &   & & 86.50  \\
  & $\surd$ &   & &86.90  \\
 &  $\surd$  &$\surd$  &  &87.63   \\
 &  $\surd$ &$\surd$ &   $\surd$&\textbf{87.85}  \\
$\surd$  &   & &   & 87.60  \\
 $\surd$  &$\surd$&   &   &   87.80\\
 $\surd$ &   $\surd$   &$\surd$   & &88.23\\
 $\surd$ &  $\surd$   &$\surd$   &$\surd$ &\textbf{88.55}\\
\bottomrule
\end{tabular}
\end{table}

\begin{table}[tp]
\center
\caption{Evaluation of all components of our RAN along with face alignment on AffectNet \textbf{without oversampling}. }
\label{tab:AffectNet}
\begin{tabular}{@{}ccccc@{}}
\toprule
 Align & Self-att. & Relation-att. & RB-Loss & Accuracy  \\
\midrule
  & &   & & 49.00  \\
 &  $\surd$  &$\surd$  &  &52.05      \\
 &  $\surd$ &$\surd$ &   $\surd$&\textbf{52.97}   \\
 $\surd$  &   & &   & 50.32  \\
 $\surd$ &   $\surd$   &$\surd$   & & 52.17\\
 $\surd$ &  $\surd$   &$\surd$   &$\surd$ & 52.50
 \\
\bottomrule
\end{tabular}
\end{table}


\begin{figure}[t]
\center
\includegraphics[width=0.95\linewidth]{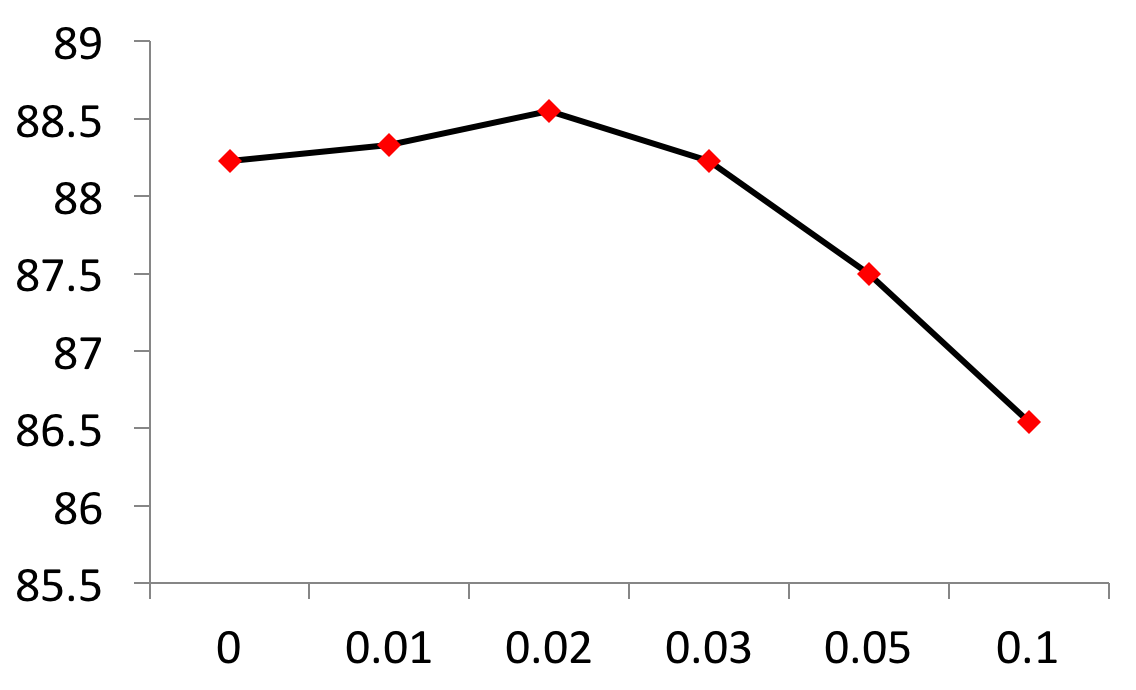}
\caption{The evaluaiton of the margin ($\alpha$) in RB-Loss on FERPlus.}
\label{fig:pbloss}
\end{figure}

\textbf{Region biased loss}. The RB-Loss is added to the self-attention module with margin 0.02 by default. From Table \ref{tab:FERPlus} and Table \ref{tab:AffectNet}, we can see that the designed RB-Loss further improves performance on both FERPlus and AffectNet consistently. Specifically, the improvement on AffectNet without face alignment is 0.92\%. With oversampling, our RAN with RB-Loss achieves \textbf{59.5\%} on the validation set of AffectNet. It is worth noting that RB-Loss does not increase computational cost in testing. 

We also evaluate the parameter $\alpha$ of RB-Loss in Figure \ref{fig:pbloss}. Increasing $\alpha$ from 0 to 0.02 gradually improves the performance while larger $\alpha$ leads to fast degradation, which indicates the original image is also important for FER. \pxj{As the mater of fact, the result of this experiment is part of our motivation to keep the original face image for our method.}

\begin{figure}[t]
\center
\includegraphics[width=1\linewidth]{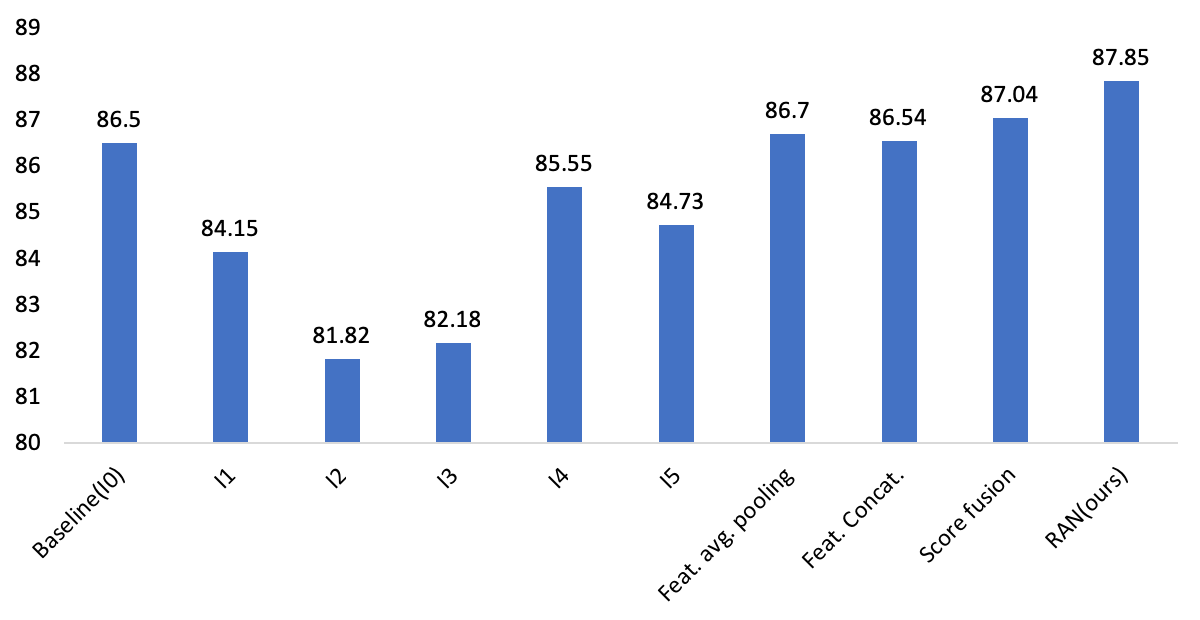}
\caption{\peng{Performce comparison of individual regions and different aggregation schemes on FERPlus.}}
\label{fig:combFERPlus}
\end{figure}

\begin{figure}[t]
    \centering
    \includegraphics[width=1\linewidth]{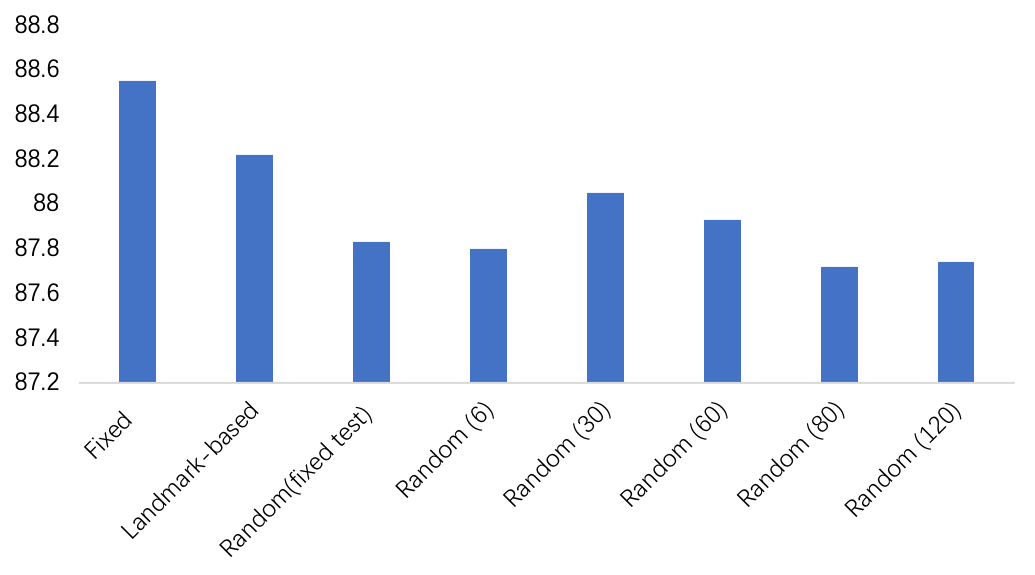}
    \caption{Evaluation of different region generation strategies and the number of random regions.}
    \label{fig:regionscheme}
\end{figure}


\textbf{Evaluation of individual regions and different fusion schemes}.
We conduct an evaluation of individual regions and different fusion schemes on the full FERPlus test datasets without face alignment. For the fusion schemes, we mainly consider three popular methods, namely feature concatanation, feature average pooling, and score fusion (i.e. score average). The evaluation results are shown in Figure \ref{fig:combFERPlus}. Several observations can be concluded as follows. First, all the individual crops are inferior to the original image which indicates the performance gain is not from special enlarged crops. Second, compared to the original image, there is no obvious improvement by concatanating and averaging region features. Third, score fusion slightly improves the baseline by 0.54\% while our RAN outperforms the baseline by 1.35\%.


\textbf{Evaluation of region generation strategies}.
We evaluate the fixed cropping, landmark-based cropping, and random cropping methods on FERPlus with the default setting for other parameters, the results are shown in Figure \ref{fig:regionscheme}. For random cropping, we randomly generate 3 regions for each image in each training iteration while generate 6, 30, 60, 80, 120 random regions for test evaluation several times. For landmark-based cropping, we set the radius as 0.4 of the side of image which ensures a similar size as fixed cropping. The fixed cropping strategy consistently outperforms the random cropping even dozens of times more regions are used. The landmark-based cropping performs slightly worse than the fixed cropping. Training with random cropping yet testing with the same fixed cropping has limited effect for random region cropping. Increasing the crops boosts performance in the beginning while degrades after 30 crops. This may be explained by that increasing crops leads to too many sub-optimized regions and they dominate the final representation.

\begin{figure}
\center
\includegraphics[width=0.8\linewidth]{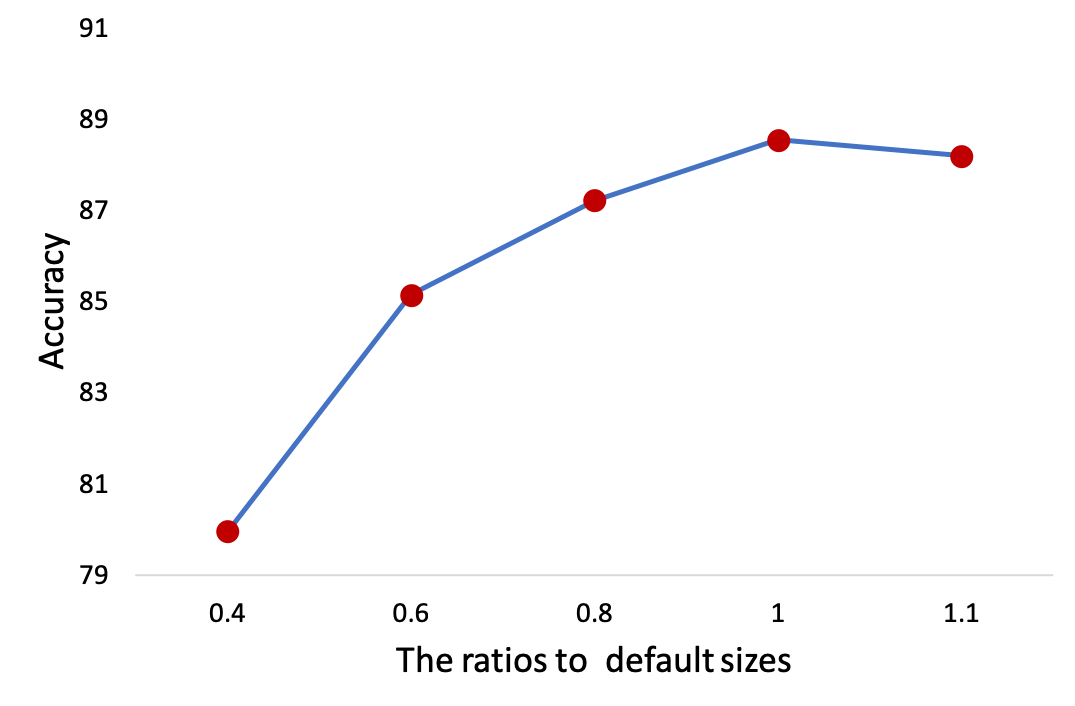}
\caption{Evaluation of region sizes on FERPlus. The ratios are compared to the default setting.}
\label{fig:sizeEval}
\end{figure}

\textbf{Evaluation of region size.} 
To explore the impact of region sizes for our RAN, we evaluate the region size of fixed cropping scheme on FERPlus with other parameter as default. Since five regions with different sizes are cropped in our default setting, we evaluate these region sizes using a ratio from low to high compared to the default sizes.The evaluation results are shown in Figure \ref{fig:sizeEval}. The performance degrades significantly with the ratio reducing to 0.4. Increasing (i.e. ration:1.1) size upon the default one slightly reduces the performance. It may be explained that the regions of $I_4$ and $I_5$ almost degrade to the original image, and the information gain from enlarging regions disappeared if too large regions are used.

\rpxj{\textbf{Evaluation of inference time.}
Since our RAN has five times feedforward operations than the baseline, we investigate the inference time on FERPlus test set.  
We evaluate the average per-image inference time and the run-time is obtained on a TITAN 1080ti GPU of Linux Cluster with a 2.6 GHz Intel(R) Xeon(R) E5-2690 CPU. The average inference time of RAN and the baseline are 0.025s and 0.006s, respectively. Due to the powerful parallel ability of GPU, the increasing time is not linear to the number of regions. 
}


\subsection{Comparison with the state-of-the-art methods}
In this section, we compare our best results to several state-of-the-art methods on FERPlus, AffectNet, SFEW, and RAF-DB.

\textbf{Comparison on FERPlus}.
We compare our RAN to several state-of-the-art methods on the FERPlus dataset in Table \ref{tab:soa_1}. Both \cite{barsoum2016training} and \cite{Albanie18} leverage the label distribution for each face as supervision. \cite{Albanie18} pretrains a SeNet50~\cite{hu2018squeeze} on VGGFace2.0~\cite{cao2018vggface2} which includes amount of large-pose faces. With the KLDiv loss and label distribution supervision, we fine-tune the public VGGFace model (VGG16 pretrained on VGGFace1.0) with our RAN and achieve 89.16\% which is a new state of the art to our knowledge.


\begin{table}[tp]
\center
\caption{Comparison to the state-of-the-art results on the FERPlus dataset.$^*$These results are trained using label distribution as supervison.}
\label{tab:soa_1}
\scriptsize
\begin{tabular}{@{}ccccc@{}}
\toprule
 Method   &Network &Pre-trained Dataset&Year & Performance  \\
 \midrule
 ~\cite{barsoum2016training}$^*$   &VGG13 & / &2016 & 85.1  \\
~\cite{huang2017combining}  &ResNet18+VGG16& / &2017 & 87.4  \\
\cite{Albanie18}$^*$   &SeNet50&VGG\_Face2&2018 & 88.8  \\
 RAN-ResNet18   &ResNet18&MS\_Celeb\_1M&2019 & 88.55  \\
 RAN-VGG16$^*$ &VGG16&VGG\_Face&2019 & \textbf{89.16}  \\
\bottomrule
\end{tabular}
\end{table}


\textbf{Comparison on AffectNet}.
Table \ref{tab:soa_2} presents the comparison on AffectNet. We obtain 52.97\% and 59.5\% without and with oversampling, respectively. It is worth noting that \cite{mollahosseini2017affectnet} only achieves 47\% with upsampling and \cite{zeng2018facial} uses one more large-scale  FER dataset and 80 layers ResNet for training with elaborated loss weights on them.

\begin{table}[tp]
\center
\caption{Comparison to the state-of-the-art results on the AffectNet dataset.$^+$Oversampling is used for a final performance report since AffectNet is imbalanced. $^\ddagger$RAF-DB is added into training data.}
\label{tab:soa_2}
\begin{tabular}{@{}cccc@{}}
\toprule
 Method   &Network &Year & Performance  \\
 \midrule
Up-Sampling~\cite{mollahosseini2017affectnet}   &AlexNet  &2018 & 47.0  \\
Weighted-Loss~\cite{mollahosseini2017affectnet}  &AlexNet &2018 &58.0  \\
\cite{zeng2018facial}$^\ddagger$   &ResNet80 &2018 & 55.71  \\
 RAN-ResNet18   &ResNet18&2019 & 52.97  \\
RAN-ResNet18$^+$ &ResNet18&2019 & \textbf{59.5}  \\
\bottomrule
\end{tabular}
\end{table}

\textbf{Comparison on SFEW}.
Table \ref{tab:soa_3} presents the comparison on SFEW. \cite{cai2018island} applies a small CNN with an island loss which is the combination of the Center loss~\cite{wen2016discriminative} and an inter-class loss. \cite{yu2015image} ensembles multiple CNNs with each CNN model initialized randomly or pretrained on FER2013. Our RAN with single model achieves 54.19\% on the validation set which is the best single model to our best of knowledge. Since model ensemble is popular on SFEW, we also conduct a naive model fusion by averaging the scores of  ResNet18 and VGG16 which obtains 56.4\%.

\begin{table}[tp]
\center
\caption{Comparison to the state-of-the-art results on the SFEW dataset.}
\label{tab:soa_3}
\scriptsize
\begin{tabular}{@{}cccc@{}}
\toprule
 Method   &Pre-trained Dataset&Year & Performance  \\
 \midrule
Island Loss~\cite{cai2018island}    &FER2013 &2018 & 52.52  \\
Identity-aware CNN~\cite{meng2017identity}   &FER2013 &2017 & 50.98  \\
Multiple deep CNNs~\cite{yu2015image}   &FER2013 &2015 & 55.96  \\
RAN-ResNet18   &MS\_Celeb\_1M&2019 & 54.19  \\
RAN(VGG16+ResNet18) &MS\_Celeb\_1M&2019 & \textbf{56.4}  \\
\bottomrule
\end{tabular}
\end{table}

\begin{table}[tp]
\center
\caption{Comparison to the state-of-the-art results on the RAF-DB dataset.}
\label{tab:soa_4}
\scriptsize
\begin{tabular}{@{}cccc@{}}
\toprule
 Method   &Network &Year & Performance  \\
 \midrule
DLP-CNN~\cite{8453893}  & 8-layer baseDCNN & 2019 & 84.13  \\
gACNN~\cite{8576656} &VGG16 & 2018 &85.07 \\
RAN-ResNet18   &ResNet18&2019 & \textbf{86.90}  \\
\bottomrule
\end{tabular}
\end{table}

\textbf{Comparison on RAF-DB}.
Table \ref{tab:soa_4} presents the comparison on RAF-DB. RAF-DB is a latest facial expression dataset which not only has basic emotion categories but also compound categories. We report the overall accuracy on the basic emotion categories.
\cite{8453893} introduces the RAF-DB dataset and uses a locality-preserving loss for network training. \cite{8576656} leverages patch-based attention networks and glocal networks. Our proposed RAN achieves \textbf{86.9\%} on RAF-DB with default setting, which are 2.77\% and 1.83\% better than DLP-CNN~\cite{8453893} and \cite{8576656}, respectively.

\section{Conclusion}
In this paper, we address the facial expression recognition in the real-world occlusion and pose-variant conditions. We build several new FER test datasets on these conditions, and propose the Region Attention Network (RAN) which adaptively adjusts the importance of facial parts. We further design a region Biased loss (RB-Loss) function to encourage high attention weight for the most important region. We evaluate our method on the collected datasets and make extensive studies on FERPlus and AffectNet. Our proposed method achieves state-of-the-art results on FERPlus, SFEW, RAF-DB, and AffectNet.


%
\IEEEpeerreviewmaketitle

\bibliographystyle{plain}

%







\end{document}